\documentclass[conference,a4paper]{IEEEtran}

\usepackage{cite}
\usepackage{amsmath,amssymb,amsfonts}
\usepackage{algorithmic}
\usepackage[pdftex]{graphicx}

\usepackage{siunitx} 
\usepackage{textcomp}
\usepackage{xcolor}
\usepackage{url}
\usepackage{svg}
\usepackage{orcidlink}

\def\BibTeX{{\rm B\kern-.05em{\sc i\kern-.025em b}\kern-.08em
    T\kern-.1667em\lower.7ex\hbox{E}\kern-.125emX}}

\begin{document}

\title{A Smart Robotic System \\ for Industrial Plant Supervision} 

\author{ 
	\IEEEauthorblockN{ 
		D. Adriana G\'omez-Rosal\IEEEauthorrefmark{1}\orcidlink{0000-0003-2373-9716},
          Max Bergau\IEEEauthorrefmark{2}\orcidlink{0009-0000-4068-8936},
          Georg K.J. Fischer\IEEEauthorrefmark{3}\orcidlink{0000-0003-1460-1061},
          Andreas Wachaja\IEEEauthorrefmark{4},
          Johannes Gr\"ater\IEEEauthorrefmark{4},
          \\
          Matthias Odenweller\IEEEauthorrefmark{5},
          Uwe Piechottka\IEEEauthorrefmark{5},
          Fabian Hoeflinger\IEEEauthorrefmark{6}\orcidlink{0000-0001-5877-1439},
          Nikhil Gosala\IEEEauthorrefmark{1}\orcidlink{0000-0003-3226-5356},
          \\
          Niklas Wetzel\IEEEauthorrefmark{1}\orcidlink{0000-0002-0502-3215},
        Daniel B\"uscher\IEEEauthorrefmark{1}\orcidlink{0000-0001-9321-3029},
        Abhinav Valada\IEEEauthorrefmark{1}\orcidlink{0000-0003-4710-3114},
		Wolfram Burgard\IEEEauthorrefmark{7}\orcidlink{0000-0002-5680-6500}
  } 
    \IEEEauthorblockA{\IEEEauthorrefmark{1}Department of Computer Science, University of Freiburg, Germany}
    \IEEEauthorblockA{\IEEEauthorrefmark{2}Sensors Automation Lab, Endress+Hauser Digital Solutions GmbH, Freiburg, Germany}
    \IEEEauthorblockA{\IEEEauthorrefmark{3}Fraunhofer Institute for Highspeed Dynamics, Ernst-Mach-Institute (EMI), Freiburg, Germany}
    \IEEEauthorblockA{\IEEEauthorrefmark{4}dotscene GmbH, Freiburg, Germany}
    \IEEEauthorblockA{\IEEEauthorrefmark{5}Evonik Operations GmbH, Essen, Germany}
    \IEEEauthorblockA{\IEEEauthorrefmark{6}Telocate GmbH, Freiburg, Germany}   
	\IEEEauthorblockA{\IEEEauthorrefmark{7}Department of Engineering, University of Technology Nuremberg, Germany}

}


\maketitle


\begin{abstract}
In today's chemical plants, human field operators perform frequent integrity checks to guarantee high safety standards, and thus are possibly the first to encounter dangerous operating conditions.
To alleviate their task,
we present a system consisting of an autonomously navigating robot integrated with various sensors and intelligent data processing. 
It is able to detect methane leaks and estimate its flow rate, detect more general gas anomalies, recognize oil films, localize sound sources and detect failure cases,
map the environment in 3D, and navigate autonomously, employing recognition and avoidance of dynamic obstacles.
We evaluate our system 
at a wastewater facility in full working conditions. Our results demonstrate that the system is able to robustly navigate the plant and provide useful information about critical operating conditions.
\end{abstract}


%
\IEEEpeerreviewmaketitle

\section{Introduction}
Chemical production plants are required to meet high safety standards.
In case of a fault, integrated safety systems are meant to automatically bring the industrial plant into a safe state.
However, they cannot detect all possible abnormal working conditions,
so, human field operators perform additional inspection rounds to check the plant's integrity.
Their tasks comprise recording local gauge values, visual and olfactory inspection for leakages of gases and fluids,
temperature inspection, and listening for unusual noises.
For these tasks, a skilled and experienced labor force is necessary, which is, in part due to demographic changes, incrementally challenging to find.
Robots equipped with a multimodal sensor setup and corresponding sensor data interpretation capabilities are envisioned to overtake such duties.

The employment of robots in dangerous conditions has a long tradition~\cite{EDWARDS198445},~\cite{5262922}. In~\cite{8722887}, swarm robots performed mobile robot tasks in a simulation with conditions of chemical leakages, radiation, and high temperatures. In~\cite{s20123461}, mobile robots search unknown radioactive sources and leakages of nuclear and chemical substances, simulating highly dangerous scenarios. While it is common to integrate mobile robots with vision and range sensors, few works with sensors for the chemical industry have been investigated, mainly due to technical difficulties~\cite{6247459}.

Recent works in the domain of deep learning
\cite{sirohi2022uncertaintyawareCam,sirohi2022uncertaintyawareLidar} show promising results with the potential for an improved understanding of the sensor data,
which is highly relevant to the autonomous supervision task.

\begin{figure}[tbp]
\centerline{\includegraphics[width=0.7\linewidth]{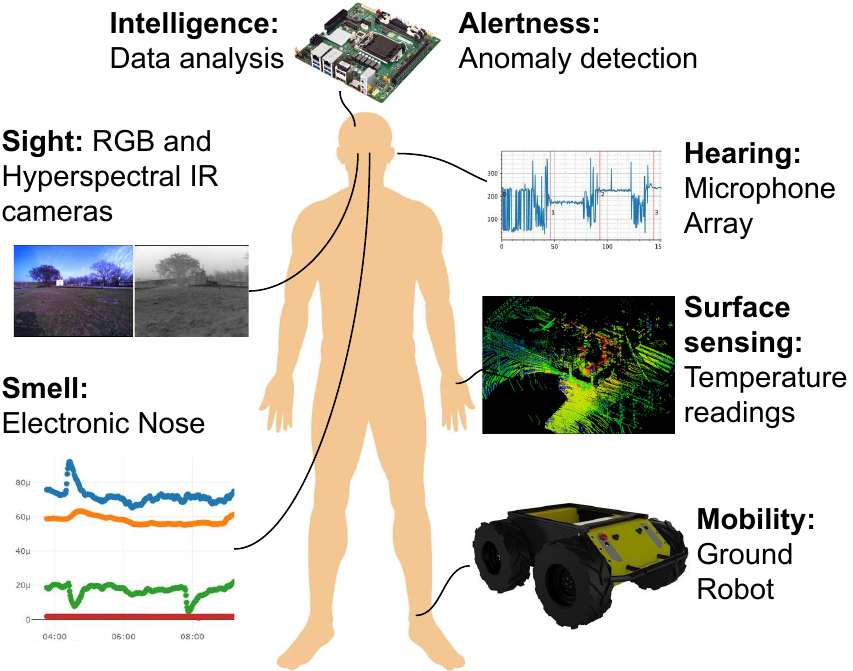}}
\caption{Our robotic system intends to replicate and extend the human senses of sight, smell and hearing for the supervision of chemical production plants.} \label{fig:isahuman}
\end{figure}

In this paper, we present an intelligent robot for autonomous supervision of industrial plants that 
aims to provide non-human automated inspection for chemical production plants.
The system consists of an autonomous navigating robot integrated with sensor modalities, that aim to resemble the human sensing capabilities of sight, smell, and hearing, as depicted in Figure~\ref{fig:isahuman}.
The strength of our system lies in the integration of diverse, novel, and complementary sensors along with data analysis based on Artificial Intelligence (AI) techniques for the automation of the surveillance task. This sets our system apart from existing ones~\cite{GoogleRob,droneplant}.



\section{System description}

\begin{figure}[tbp]
\centerline{\includegraphics[width=0.6\linewidth]{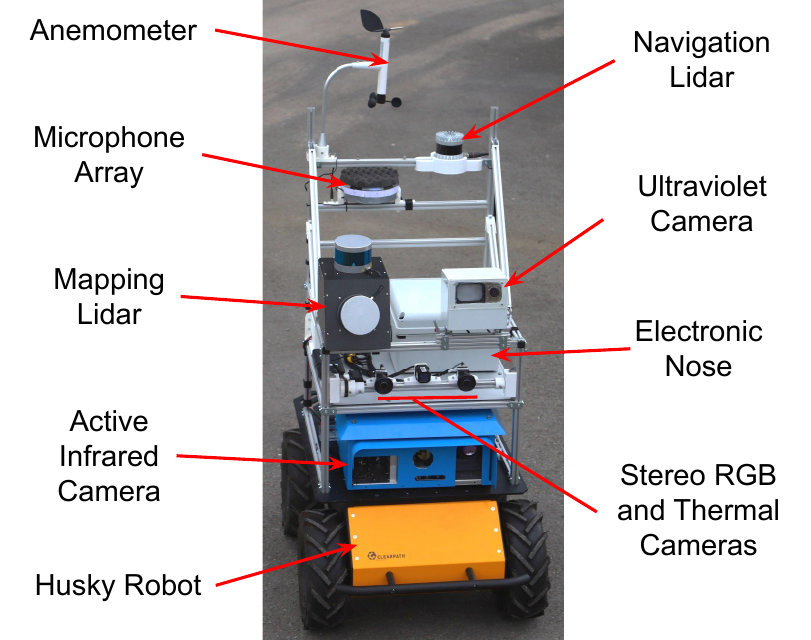}}
\caption{Our robotic system is based on the Husky mobile platform and is integrated with diverse sensors.} \label{fig:robot}
\end{figure}

Our robotic system, as shown in Figure~\ref{fig:robot}, consists of a mobile robot platform integrated with diverse sensors.
The robot navigates the industrial plant autonomously with its onboard processing capabilities.
However, tasks that require more resources, such as object or anomaly detection, are executed on a remote server.
To this end, the data is transmitted over the wireless network employing the Robot Operating System (ROS)~\cite{ROS}.

\subsection{Mobile robot}

We employ the Clearpath Husky A200 as the mobile platform since it is a rugged,
all-terrain robot equipped with an onboard computer that is designed to operate in indoor and outdoor environments.
The sensor arrangement weighs about 35 kg and is provided with power by the Husky base.

\subsection{Electronic nose}

For the task of gas anomaly detection, we developed an \textit{Electronic Nose} (E-Nose).
Its working principle is based on the integration of different gas sensors.
We employ three non-selective Metal Oxide Semiconductor (MOX) sensors and,
in combination with optical (non-dispersive infrared, laser scattering) technologies,
we measure air contaminants, methane, CO2, and flammable gases.
The second category of gas sensors, based on electrochemical principles, are specific and fast in their response.
This data allowed the training of a supervised learning network able to identify gas signatures.
Additionally, relevant information, such as humidity and wind speed and direction, was logged.

\subsection{Active infrared gas camera}

We developed an active infrared gas camera (IRcam) for the detection of methane leaks, and the estimation of gas concentration and flow rate.
The light detection is carried out by an ImageIR 8300 camera from Infratec,
while its field of view is illuminated by a 25 mW tunable $\sim$\SI{3260}{\nano\meter} single mode DFB-interband cascade laser from nanoplus.
Three consecutive images are recorded while the lasers' wavelength is tuned over a methane absorption line.
We process the image batches as described in~\cite{bergau_real-time_2023} to obtain concentration length information.
To ensure wavelength stability, a reference gas cell is filled with methane.
We feed the gas-image stream at 5 Hz to a 3D convolutional neural network (CNN) trained to estimate the flow rate of the gas leak in realtime.

\subsection{Ultraviolet camera}

We developed a remote fluorescence detection system for the recognition of oil films on surfaces and potential oil rests.
This system consists of a \SI{365}{\nm} Ultraviolet (UV) LED (ODS75 Smart Vision Lights) and a CMOS camera (OpenMV Cam H7R2). 
Pairs of images are taken by the camera, one image with ambient light and one with additional UV illumination.
The difference between these two frames yields an image containing UV-excited fluorescence,
to reveal information about oil films due to their aromatic molecular structures~\cite{stasiuk97}.
While this sensor is able to detect oil films up to a distance of \SI{1}{\m} in indoor lighting conditions,
the sensitivity decreases outdoors, in particular in sunny conditions.

\subsection{Microphone array}

To provide the robot with hearing capabilities, we developed a five-microphone uniform circular array.
We employ TDK ICS-40720 MEMS-microphones with a sampling rate of \SI{96}{\kilo\hertz},
and an inter-microphone spacing of $\approx\, $\SI{8}{\centi\meter}.
The setup is capable to localize sound sources with a maximum frequency of around \SI{2.1}{\kilo\hertz}
using a specifically designed method for direction-of-arrival estimation\cite{9662589}.
Our algorithm is able to directly resolve distinct sources,
even though only a few microphones are available in our setup, constrained in size.

We detect sound anomalies by recognizing sound samples as normal or abnormal through a supervised learning model.
The network is trained on the MIMII~\cite{purohit2019mimii} dataset
and can be extended to include sound samples from the plant where the robot is operated.
The anomalies, constituted by unknown sound samples, are detected using a deep neural network with an encoder-decoder architecture.
To this end, the network learns features of the sound samples by encoding the sample into a latent vector embedding
and then, decoding the vector to recreate the input sound sample.
For unknown sounds, the input cannot be reconstructed, which indicates an anomaly.

\subsection{Mapping LiDAR setup}

We developed a task-specific LiDAR setup  to build  accurate 3D maps of the environment of the robot. We furthermore map the deep sewage channels at the plant.
The unit comprises two Velodyne VLP16 laser scanners mounted on a cuboid structure.
While the top laser is leveled with the ground to capture the distant surroundings,
the front laser scanner is steeply inclined, allowing to capture the ground and possible channels in front of the robot.
We employ an Inertial Measurement Unit (IMU) to track the short-term motion of the system, increasing mapping accuracy and robustness.
A dense 3D map of the environment is obtained by fusing the continuously captured data with a 3D 
Simultaneous Localization and Mapping (SLAM) approach when the robot is in motion.
The resulting map of the environment is represented as a point cloud in 3D space.
With this generic representation, various applications are realized, such as embedding other sensor information (like temperature),
geometric alignment for inventory purposes, changes or anomalies detection, and lastly semantic segmentation with deep learning techniques.

\subsection{Passive cameras}

The optical perception system includes three forward-facing cameras composed of
a stereo rig of two RGB cameras, Blackfly S GigE FLIR, with a resolution of 1544x2064, 
and a thermal camera, FLIR Boson Long wave infrared (LWIR), with 512x640 pixels
and a spectral range of \SI{7.5}{\micro\meter}-\SI{13.5}{\micro\meter}.
These three cameras are synchronized with an external trigger at a frame rate of 30 fps. 

The live object detection module employs the images from the RGB cameras to support the navigation functions.
It detects dynamic agents in the scene, such as cars and pedestrians, and is based on CenterNet~\cite{centernet}, a
network for object detection.
It is a fully-supervised model that first predicts the object centers and their corresponding center offsets,
and then combines these values to estimate bounding boxes around dynamic objects. 
To overcome the problem of data labeling, we leveraged existing object detection ground truth data from the KITTI~\cite{geiger2013vision} autonomous driving dataset and
then adapted the model by fine-tuning it on 150 manually annotated images from our chemical plant.

\subsection{Navigation LiDAR}

To enable autonomous navigation, we use an OS128 Ouster LiDAR.
It has a vertical field of view of 90\textdegree{}, 50 m range, 128 times 1024 channels at a 20 Hz rate, and a built-in IMU.
The navigation functionality consists of the tasks of localization, mapping, and path planning.
While the localization approach in a known 2D grid map builds on Adaptive Monte Carlo Localization (AMCL)~\cite{probabilisticrobotics},
using a Rao-Blackwellized particle filter (RBPF), we solved the SLAM task with an efficient RBPF that creates grid maps~\cite{OpenslamGmapping}.
To this end, the 3D point clouds from the Ouster LiDAR were projected into 2D range scans.

The path planning task involves two hierarchized planners.
While we solve the global planning using the $A^*$~\cite{Astar} approach, we employ the more advanced timed elastic bands (TEB)~\cite{TEB} approach 
for local planning. It is complemented by a reinforcement learning (RL) algorithm supporting the avoidance of dynamic obstacles.
The RL algorithm uses the dynamic obstacles detected in the RGB images and projects them in the birds-eye view (BEV) using the Ouster LiDAR data.
We combine features extracted from this local semantic map via a CNN with the robot's low-dimensional internal state,
such as the navigation goal and recent controls.
Several RL agents were trained using the Soft Actor-Critic~\cite{softactorcritic} algorithm with different hyperparameters, to compare the ability to predicatively navigate around dynamic objects in simulations and real-world scenarios.

\section{Experiments}

\begin{figure}[tbp]
\centerline{\includegraphics[width=1.0\linewidth]{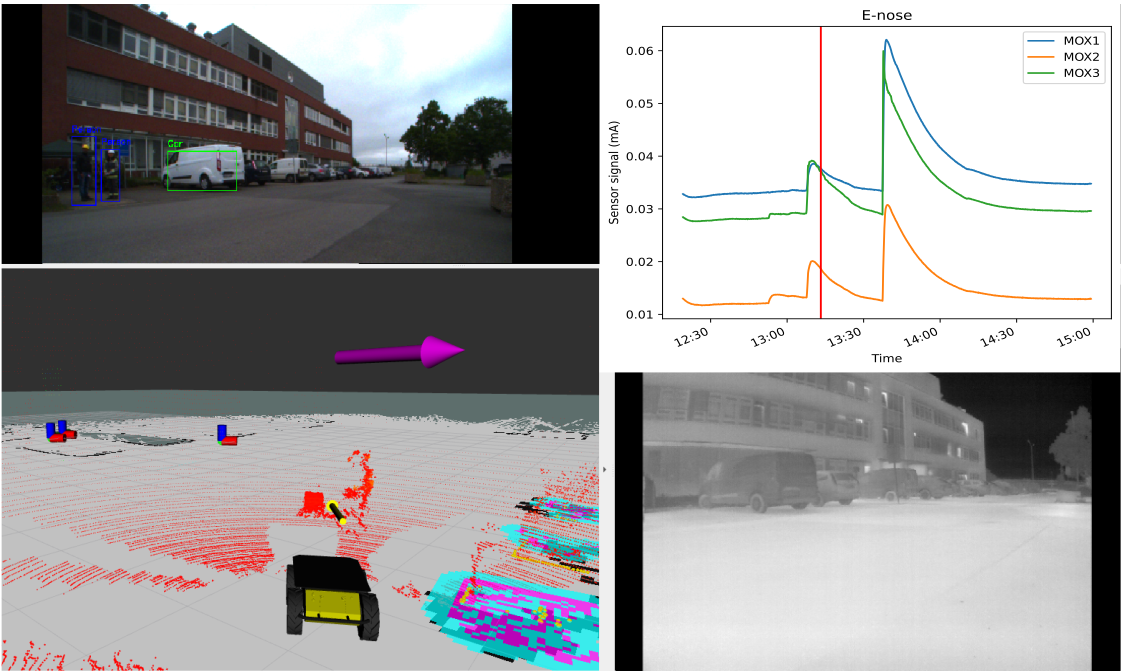}}
\caption{Visualization of the sensor data.
Upper left: RGB image with live dynamic object detection.
Lower left: robot localization in the 2D occupancy grid map (gray),
with the projected dynamic objects (red/blue),
navigation LiDAR data (red),
static obstacles (cyan/pink),
wind speed and direction (purple arrow),
and audio noise direction (yellow arrow).
Upper right: E-Nose data (red line: readings at the other data's timestamp).
Lower right: thermal image.
} \label{fig:rviz}
\end{figure}

We chose the wastewater area of the Marl chemical plant in Germany for on-site tests. This area allows for a ground-level route along various points of interest including
hot pipes, loud pumps, and sewage canals with occasionally increased gas levels.
In addition, samples of silicon oil, vinegar, engine oil, and an artificial methane leak
were placed along the inspection path to test the robot's capabilities.
Figure~\ref{fig:rviz} shows a sample of the live robot state and sensor data transmitted to the remote computer,
including the detection of dynamic objects, their projection into the BEV, further robot navigation data, the thermal image, and a longer sample from the MOX gas sensors.

Tests using the IRCam allowed the live visual detection and quantification of synthetic methane leaks in the plant
with a gas flow rate of 20 to 200 mL/min from a 2 m distance with static measurements. 
We also tested oil detection with the UV camera
by placing mineral oil on the metal surface of a pipe partially covered by sunlight. The sample was recognized at an approximate distance of 50 cm.

Using the microphone array, multiple pump noises were detected with even small angular distances between two pumps being discriminated.
The direction of the strongest sound source was estimated with a median error of around \SI{5}{\degree} in azimuthal direction at rest.
Various motor vehicles, such as cars and trucks, and pumps with artificial noise sources were identified by sound pattern recognition.

We employed the mapping LiDAR to extract 3D information about the environment and for embedding additional sensor data,
e.g., images from the thermal camera were mapped and visualized in the 3D model.
We identified geometric changes in the environment between inspection runs, and 
demonstrated an online 3D mapping capability.
In conclusion, we successfully tested our system for its capability to autonomously perform tasks of chemical plant inspection.

\clearpage  

\bibliographystyle{IEEEtran}
\bibliography{references.bib}

\begin{thebibliography}{10}
\providecommand{\url}[1]{#1}
\csname url@samestyle\endcsname
\providecommand{\newblock}{\relax}
\providecommand{\bibinfo}[2]{#2}
\providecommand{\BIBentrySTDinterwordspacing}{\spaceskip=0pt\relax}
\providecommand{\BIBentryALTinterwordstretchfactor}{4}
\providecommand{\BIBentryALTinterwordspacing}{\spaceskip=\fontdimen2\font plus
\BIBentryALTinterwordstretchfactor\fontdimen3\font minus
  \fontdimen4\font\relax}
\providecommand{\BIBforeignlanguage}[2]{{%
\expandafter\ifx\csname l@#1\endcsname\relax
\typeout{** WARNING: IEEEtran.bst: No hyphenation pattern has been}%
\typeout{** loaded for the language `#1'. Using the pattern for}%
\typeout{** the default language instead.}%
\else
\language=\csname l@#1\endcsname
\fi
#2}}
\providecommand{\BIBdecl}{\relax}
\BIBdecl

\bibitem{EDWARDS198445}
\BIBentryALTinterwordspacing
M.~Edwards, ``Robots in industry: An overview,'' \emph{Applied Ergonomics},
  vol.~15, no.~1, pp. 45--53, 1984. [Online]. Available:
  \url{https://www.sciencedirect.com/science/article/pii/S0003687084901212}
\BIBentrySTDinterwordspacing

\bibitem{5262922}
F.~He, Y.~Sun, X.~liu, and Z.~Du, ``Humanoid detection of indoor dangerous gas
  source by mobile robot,'' in \emph{2009 IEEE International Conference on
  Automation and Logistics}, 2009, pp. 237--242.

\bibitem{8722887}
X.~Huang, F.~Arvin, C.~West, S.~Watson, and B.~Lennox, ``Exploration in extreme
  environments with swarm robotic system,'' in \emph{2019 IEEE International
  Conference on Mechatronics (ICM)}, vol.~1, 2019, pp. 193--198.

\bibitem{s20123461}
\BIBentryALTinterwordspacing
J.~Huo, M.~Liu, K.~A. Neusypin, H.~Liu, M.~Guo, and Y.~Xiao, ``Autonomous
  search of radioactive sources through mobile robots,'' \emph{Sensors},
  vol.~20, no.~12, 2020. [Online]. Available:
  \url{https://www.mdpi.com/1424-8220/20/12/3461}
\BIBentrySTDinterwordspacing

\bibitem{6247459}
H.~Ishida, Y.~Wada, and H.~Matsukura, ``Chemical sensing in robotic
  applications: A review,'' \emph{IEEE Sensors Journal}, vol.~12, no.~11, pp.
  3163--3173, 2012.

\bibitem{sirohi2022uncertaintyawareCam}
K.~Sirohi, S.~Marvi, D.~Büscher, and W.~Burgard, ``Uncertainty-aware panoptic
  segmentation,'' \emph{IEEE Robotics and Automation Letters}, vol.~8, no.~5,
  pp. 2629--2636, 2023.

\bibitem{sirohi2022uncertaintyawareLidar}
K.~Sirohi, S.~Marvi, D.~B{\"u}scher, and W.~Burgard, ``Uncertainty-aware lidar
  panoptic segmentation,'' \emph{arXiv preprint arXiv:2210.04472}, 2022.

\bibitem{GoogleRob}
S.~Chakraborty, S.~Mukherjee, S.~K. Saha, and H.~N. Saha, ``Autonomous vehicle
  for industrial supervision based on google assistant services \& iot
  analytics,'' pp. 1061--1070, 2019.

\bibitem{droneplant}
\BIBentryALTinterwordspacing
F.~J. Perez-Grau, J.~R. {Martinez-de Dios}, J.~L. Paneque, J.~J. Acevedo,
  A.~Torres-González, A.~Viguria, J.~R. Astorga, and A.~Ollero, ``Introducing
  autonomous aerial robots in industrial manufacturing,'' \emph{Journal of
  Manufacturing Systems}, vol.~60, pp. 312--324, 2021. [Online]. Available:
  \url{https://www.sciencedirect.com/science/article/pii/S0278612521001321}
\BIBentrySTDinterwordspacing

\bibitem{ROS}
M.~Quigley, K.~Conley, B.~Gerkey, J.~Faust, T.~Foote, J.~Leibs, R.~Wheeler,
  A.~Y. Ng \emph{et~al.}, ``Ros: an open-source robot operating system,''
  vol.~3, no. 3.2, p.~5, 2009.

\bibitem{bergau_real-time_2023}
\BIBentryALTinterwordspacing
M.~Bergau, T.~Strahl, B.~Scherer, and J.~Wöllenstein,
  ``\BIBforeignlanguage{English}{Real-time active-gas imaging of small gas
  leaks},'' \emph{\BIBforeignlanguage{English}{Journal of Sensors and Sensor
  Systems}}, vol.~12, no.~1, pp. 61--68, Feb. 2023, publisher: Copernicus GmbH.
  [Online]. Available: \url{https://jsss.copernicus.org/articles/12/61/2023/}
\BIBentrySTDinterwordspacing

\bibitem{stasiuk97}
L.~Stasiuk and L.~Snowdon, ``Fluorescence micro-spectrometry of synthetic and
  natural hydrocarbon fluid inclusions: crude oil chemistry, density and
  application to petroleum migration,'' \emph{Applied Geochemistry}, 1997.

\bibitem{9662589}
G.~Fischer, F.~Zeqiri, A.~Gabbrielli, D.~Jan~Schott, J.~Bordoy, W.~Xiong,
  F.~Höflinger, J.~Wendeberg, K.~Fischer, C.~Schindelhauer, and
  S.~Johann~Rupitsch, ``Localization of acoustic gas leakage sources with a
  circular microphone array,'' in \emph{2021 International Conference on Indoor
  Positioning and Indoor Navigation (IPIN)}, 2021, pp. 1--7.

\bibitem{purohit2019mimii}
H.~Purohit, R.~Tanabe, K.~Ichige, T.~Endo, Y.~Nikaido, K.~Suefusa, and
  Y.~Kawaguchi, ``Mimii dataset: Sound dataset for malfunctioning industrial
  machine investigation and inspection,'' 2019.

\bibitem{centernet}
K.~Duan, S.~Bai, L.~Xie, H.~Qi, Q.~Huang, and Q.~Tian, ``Centernet: Keypoint
  triplets for object detection,'' 2019.

\bibitem{geiger2013vision}
A.~Geiger, P.~Lenz, C.~Stiller, and R.~Urtasun, ``Vision meets robotics: The
  kitti dataset,'' \emph{The International Journal of Robotics Research},
  vol.~32, no.~11, pp. 1231--1237, 2013.

\bibitem{probabilisticrobotics}
S.~Thrun, W.~Burgard, and D.~Fox, ``Probabilistic robotics (intelligent
  robotics and autonomous agents),'' \emph{The MIT Press}, 2005.

\bibitem{OpenslamGmapping}
G.~Grisetti, C.~Stachniss, and W.~Burgard, ``Improving grid-based slam with
  rao-blackwellized particle filters by adaptive proposals and selective
  resampling,'' \emph{Proc. of the IEEE International Conference on Robotics
  and Automation (ICRA)}, 2005.

\bibitem{Astar}
\BIBentryALTinterwordspacing
P.~Hart, N.~Nilsson, and B.~Raphael, ``A formal basis for the heuristic
  determination of minimum cost paths,'' \emph{{IEEE} Transactions on Systems
  Science and Cybernetics}, vol.~4, no.~2, pp. 100--107, 1968. [Online].
  Available: \url{https://doi.org/10.1109/tssc.1968.300136}
\BIBentrySTDinterwordspacing

\bibitem{TEB}
B.~Magyar, N.~Tsiogkas, J.~Deray, S.~Pfeiffer, and D.~Lane, ``Time-elastic
  bands for manipulation motion planning,'' \emph{IEEE Robotics and Automation
  Letters}, 2019.

\bibitem{softactorcritic}
T.~Haarnoja, A.~Zhou, P.~Abbeel, and S.~Levine, ``Soft actor-critic: Off-policy
  maximum entropy deep reinforcement learning with a stochastic actor,'' 2018.

\end{thebibliography}

\end{document}